\documentclass[10pt,twocolumn,letterpaper]{article}

\usepackage{spconf}
\usepackage{times}
\usepackage{epsfig}
\usepackage{amsmath,amsfonts}
\usepackage{mathtools}
\usepackage{amssymb}
\usepackage{booktabs}
\usepackage{caption}
\usepackage{url}
\usepackage{bm}
\usepackage{xfrac}
\usepackage{subfigure}
\usepackage{tabularx}
\usepackage{array}
\usepackage{float}
\usepackage{multirow,textcomp,subfigure,multirow,xspace,arydshln,cite}
\usepackage{color}

\newcommand{\Pairs}{\mathcal{P}}
\newcommand{\dom}{\mathcal{X}}
\newcommand{\RR}{\mathbb{R}}
\newcommand{\EE}{\mathbb{E}}

\newcommand{\loss}{L(\Theta|\, \Pairs)}
\newcommand{\error}{\text{error}(\alpha, \beta)}

\newcommand{\pl}{p^\text{{\tiny local}}}
\newcommand{\pg}{p^\text{{\tiny global}}}
\newcommand{\pe}{p}

\newcommand{\Cl}{C^\text{{\tiny local}}}
\newcommand{\Cg}{C^\text{{\tiny global}}}

\newcommand{\x}{\bm{x}}

\newcommand{\Pe}{\Pi}
\renewcommand{\P}{\bm{\Pe}}

\newcommand{\se}{\pi}
\newcommand{\s}{\bm{\se}}

\newcommand{\we}{\omega}
\newcommand{\w}{\bm{\we}}

\DeclareGraphicsExtensions{.eps,.pdf,.jpeg,.png}

\usepackage[breaklinks=true,bookmarks=false]{hyperref}

\title{Rank-smoothed Pairwise Learning in Perceptual Quality Assessment}

\name{Hossein Talebi, Ehsan Amid\sthanks{Ehsan Amid was partially supported by the NSF grant IIS 1546459.}, Peyman Milanfar, and Manfred K. Warmuth}
\address{Google Research \\ Mountain View, CA}
\begin{document}
\maketitle
%

\begin{abstract}
   Conducting pairwise comparisons is a widely used approach in curating human perceptual preference data. Typically raters are instructed to make their choices according to a specific set of rules that address certain dimensions of image quality and aesthetics. The outcome of this process is a dataset of sampled image pairs with their associated empirical preference probabilities. Training a model on these pairwise preferences is a common deep learning approach. However, optimizing by gradient descent through mini-batch learning means that the ``global'' ranking of the images is not explicitly taken into account. In other words, each step of the gradient descent relies only on a limited number of pairwise comparisons. In this work, we demonstrate that regularizing the pairwise empirical probabilities with aggregated rankwise probabilities leads to a more reliable training loss. We show that training a deep image quality assessment model with our rank-smoothed loss consistently improves the accuracy of predicting human preferences.
\end{abstract}

\section{Introduction}

Perceptual image quality assessment is an integral part of any imaging application. Although modeling human perceptual preferences is a challenging task, recent advances in deep learning with convolutional neural networks (CNNs) has resulted in accurate quality prediction methods. Typically, these methods rely on human perceptual ratings as ground truth labels, and train a CNN to predict the human perceptual preferences \cite{kim2017deep, talebi2018nima}. Generically, these labels are obtained from subjective studies, where human raters are asked to evaluate a single image \cite{murray2012ava, ghadiyaram2015massive} (a.k.a single stimulus), or perform a pairwise comparison \cite{ponomarenko2009tid2008, ponomarenko2015image}.

Mantiuk et al.~\cite{mantiuk2012comparison} studied four common methods for subjective image quality assessment, and concluded that the forced-choice pairwise comparison is the most accurate and efficient method. This conclusion was drawn based on analysis of variance and statistical testing. In a forced-choice pairwise comparison study, human raters are asked to compare two images based on a specific quality component (blur, noise, exposure, compression artifacts, composition, etc.), or overall aesthetics and beauty. On the downside, performing forced-choice pairwise comparisons requires a large number of trials. To be exact, for $N$ images, all possible pair combinations amounts to $N(N-1)/2$. In practice, when $N$ is large, only a small fraction of all possible pairs are rated. As shown by Silverstein et al. \cite{silverstein2001efficient}, by sorting the data with a pilot method before performing pairwise comparison study, fewer pairs are needed to reliably rank the data. 
However, the main shortcoming of this approach is that a pilot quality predictor is necessary to sort the data prior to conducting the user study.

\begin{figure}[t!]
\vspace{-2 mm}
\begin{center}
\includegraphics*[scale=0.11]{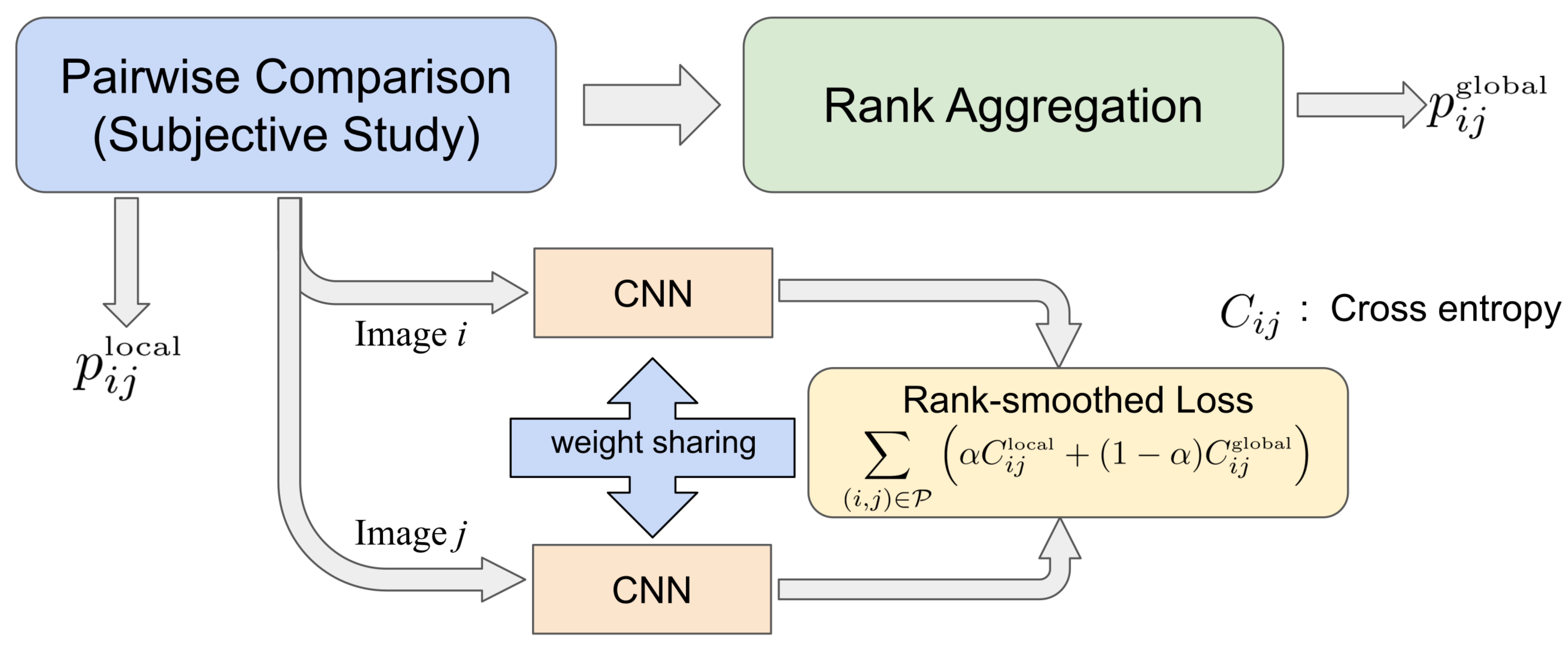}
\end{center}
\vspace{-6 mm}
{\caption{\small Proposed framework for learning from pairwise comparisons. The subjective study results in empirical preferences $\pl_{ij}$ that represent preference of image $i$ over image $j$. We use rank aggregation method to approximate global probabilities $\pg_{ij}$ that take into account the overall ranking of image $i$ and image $j$. We propose to train a CNN with a combo cross-entropy loss.} \label{fig:framework}}
\vspace{-4 mm}
\end{figure}

Ranking items after obtaining their pairwise comparisons has been studied extensively \cite{dykstra1960rank,huang2008ranking,negahban2012}. In addition to finding the ranking, obtaining a score (as intensity of the preference) for each item can be useful in various applications. Huang et al. \cite{huang2008ranking} solve a convex minimization problem for their group comparison results. More recently, Negahban et al. \cite{negahban2012} propose an algorithm (called \textit{Rank Centrality}) with a random walk interpretation over
the graph of items with edges between compared items. It turns out that the rank scores are the stationary probability of this random walk. 

Recent quality prediction approaches based on pairwise comparisons use mini-batches of image pairs and gradually learn the global scoring via gradient descent~\cite{ma2017dipiq}. Yet, this framework neglects the global ordering of the images during training. In this paper we take advantage of both pairwise (local) and rankwise (global) preferences to train a perceptual quality assessment model. Our framework is shown in Fig.~\ref{fig:framework}. Our training data consists of pairwise comparisons of images. 
We use the weight sharing technique to train a CNN~\cite{talebi2018nima, talebi2018learned} with a novel combo loss, which we call \textit{rank-smoothed} loss. Our loss combines the local preferences as well as the global ones obtained via \textit{Rank Centrality}~\cite{negahban2012} in an elegant manner, and leads to a more improved model than just using one of the preferences. In the following, we first describe learning from pairwise comparisons, and then explain our proposed method.


\section{Learning from pairwise comparisons}

Given $N$ items, the pairwise comparions are generally collected via human evaluators on a subset of all possible pairs of items, denoted by $\Pairs$. For a pair $(i,j) \in \Pairs$, the pairwise comparison is repeated multiple times across different evaluators and the aggregate information is given in the form $(n_{ij}, n_{ji})$ where $n_{ij}$ denotes the total number of times that item $i$ is preferred over item $j$. This aggregate information is then used to define 
\begin{equation}
\label{eq:p_local}
\pl_{ij} = \frac{n_{ij}}{n_{ij} + n_{ji}}\, ,
\end{equation}
which corresponds to the maximum-likelihood estimate of the Bernoulli random variable that picks item $i$ over $j$. Note that this probability is independent of all the remaining items $k \in [N], \, k \neq i, j$.

RankNet~\cite{burges2005} is perhaps the most commonly used approach for learning to rank from pairwise comparisons. The idea behind RankNet is to train a network to extract a better representation for the items that reflects the information provided by the pairwise comparisons. More technically, let $\dom \in \RR^d$ denote the domain of the input items and let $\x_i, \x_j \in \dom$ denote the input representation for the pair of items $(i,j) \in \Pairs$. In RankNet, a network $f_\Theta:\, \dom \rightarrow \RR$ (parameterized by $\Theta$) is used to extract the \emph{scores} $s_i = f_\Theta(\x_i)$ and $s_j = f_\Theta(\x_j)$ for items $i$ and $j$, respectively. Next, the probability of preferring $i$ over $j$ is defined as
\begin{equation}
\label{eq:q}
q_{ij} \coloneqq \frac{\exp(s_i-s_j)}{1 + \exp(s_i - s_j)} = \frac{\exp(s_i)}{\exp(s_i) + \exp(s_j)}\, .
\end{equation}
Thus, $s_i, s_j \in \RR$ can be viewed as logits in a binary logistic regression problem and $s_i > s_j$ indicates that item $i$ is preferred over item $j$ more often, resulting in a higher $q_{ij}$ probability.

The predicted preference probability $q_{ij}$ is then compared to the empirical probability $\pl_{ij}$ via the cross entropy loss,
$
\Cl_{ij} \coloneqq -\pl_{ij}\log (q_{ij}) - (1-\pl_{ij})\log(1 - q_{ij})\, ,
$
and the total loss of the training is defined as sum of the cross entropy losses over the set of pairwise comparisons,
\begin{equation}
\label{eq:loss-pairwise}
\loss \coloneqq \sum_{(i,j) \in \Pairs}\, \Cl_{ij}\, .
\end{equation}

Although RankNet~\cite{burges2005} can be applied to complete or incomplete comparisons (where $\pl_{ij}$ is not available for every $i,j$ pair), its main drawback is that at each step of the gradient descent when minimizing $\loss$, only a mini-batch of pairwise comparisons are taken into account. This may lead to sub-optimal learning solutions, especially in the presence of noise in the data-collection process. We show that regularizing (or smoothing) $\loss$ with an approximation of the global ranking significantly improves this shortcoming. Next, we discuss the \textit{Rank Centrality} method for approximating the global ranking.

\subsection{Proposed method}

In order to incorporate the global ordering information of the items into the learning problem, we propose minimizing the following objective function instead:
\begin{equation}
\label{eq:loss-combined}
\loss \coloneqq \sum_{(i,j)\in \Pairs} \Big(\alpha\, \Cl_{ij} + (1 - \alpha) \, \Cg_{ij}\Big)\, ,
\end{equation}
where\, $\Cg_{ij} \coloneqq -\pg_{ij}\log (q_{ij}) - (1-\pg_{ij})\log(1 - q_{ij})$ is the cross entropy divergence between the global pairwise comparison probability $\pg_{ij}$ (defined later) and $q_{ij}$. The parameter $0\leq \alpha \leq 1$ controls the trade-off between the local (empirical) loss and the global loss and the choice of $\alpha = 1$ reduces to the empirical pairwise comparison loss~\eqref{eq:loss-pairwise}. The probability $\pg_{ij}$ should effectively reflect the pairwise preference of $i$ over $j$ while maintaining information about the ordering of \emph{all} the remaining items. Note that now the minimizer of the loss~\eqref{eq:loss-combined} over the predicted preference probabilities corresponds to 
\begin{equation}
\label{eq:q-star}
q^\star_{ij} = \alpha\, \pl_{ij} + (1- \alpha)\, \pg_{ij}\,,\,\,\, (i,j)\in \Pairs\, .
\end{equation}
In order to aggregate the information provided in the pairwise comparisons between the pairs of items $\Pairs$ into a single global ordering among all items, we explore the idea of \textit{Rank Centrality} proposed in~\cite{negahban2012}.

\begin{figure*}[t!]
\vspace{-8mm}
\setlength{\tabcolsep}{4pt}
\begin{center}
\resizebox{\textwidth}{!}{
\begin{tabular}{cccc}
    {\footnotesize Effect of $n_t$ on $\alpha$}& {\footnotesize Effect of $n_t$ on $\beta$} & {\footnotesize Effect of $r$ on $\alpha$} & {\footnotesize Effect of $r$ on $\beta$}\\[-1mm]
    \subfigure{\includegraphics[ width=0.24\textwidth]{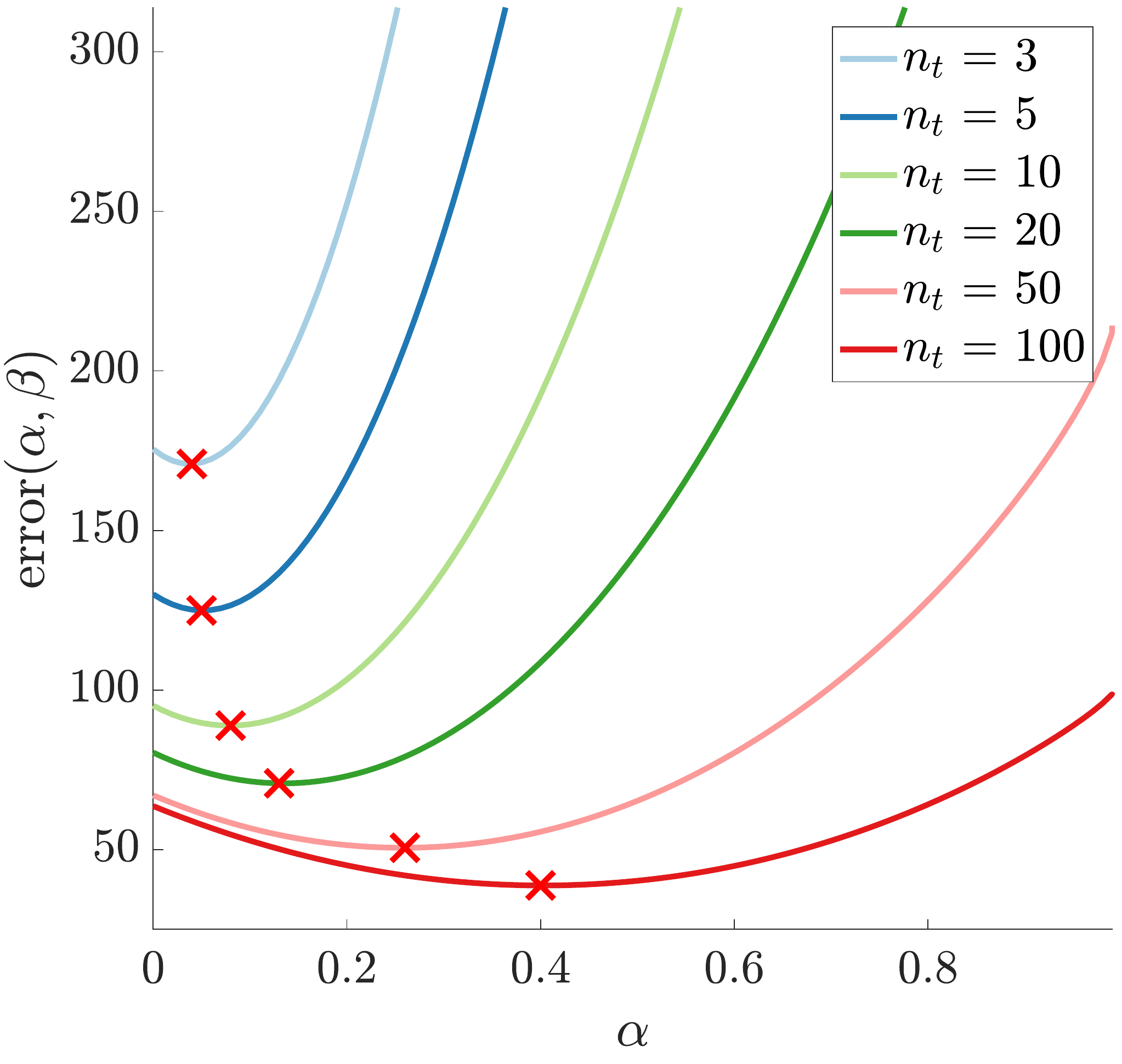}} &
    \subfigure{\includegraphics[ width=0.24\textwidth]{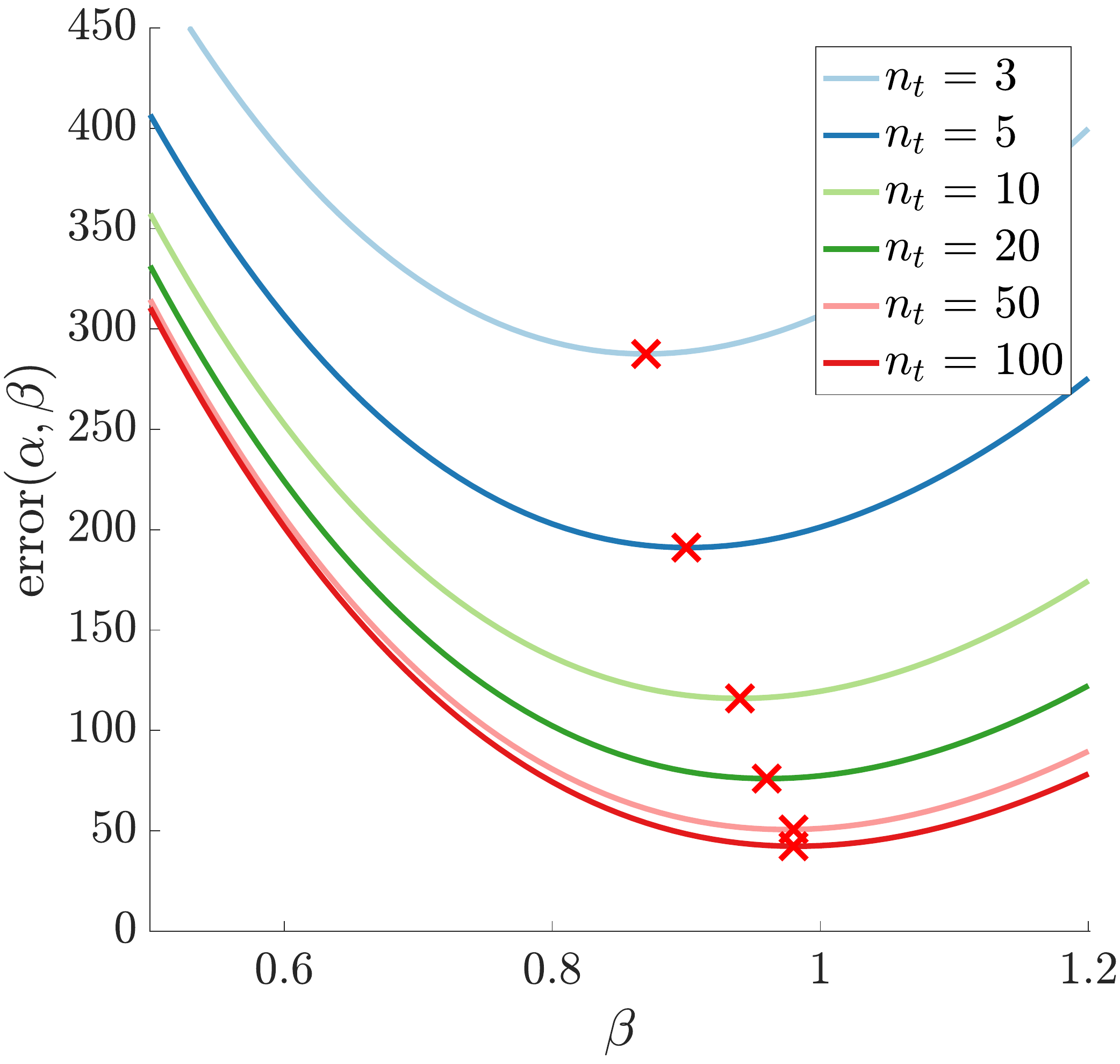}} &
    \subfigure{\includegraphics[ width=0.24\textwidth]{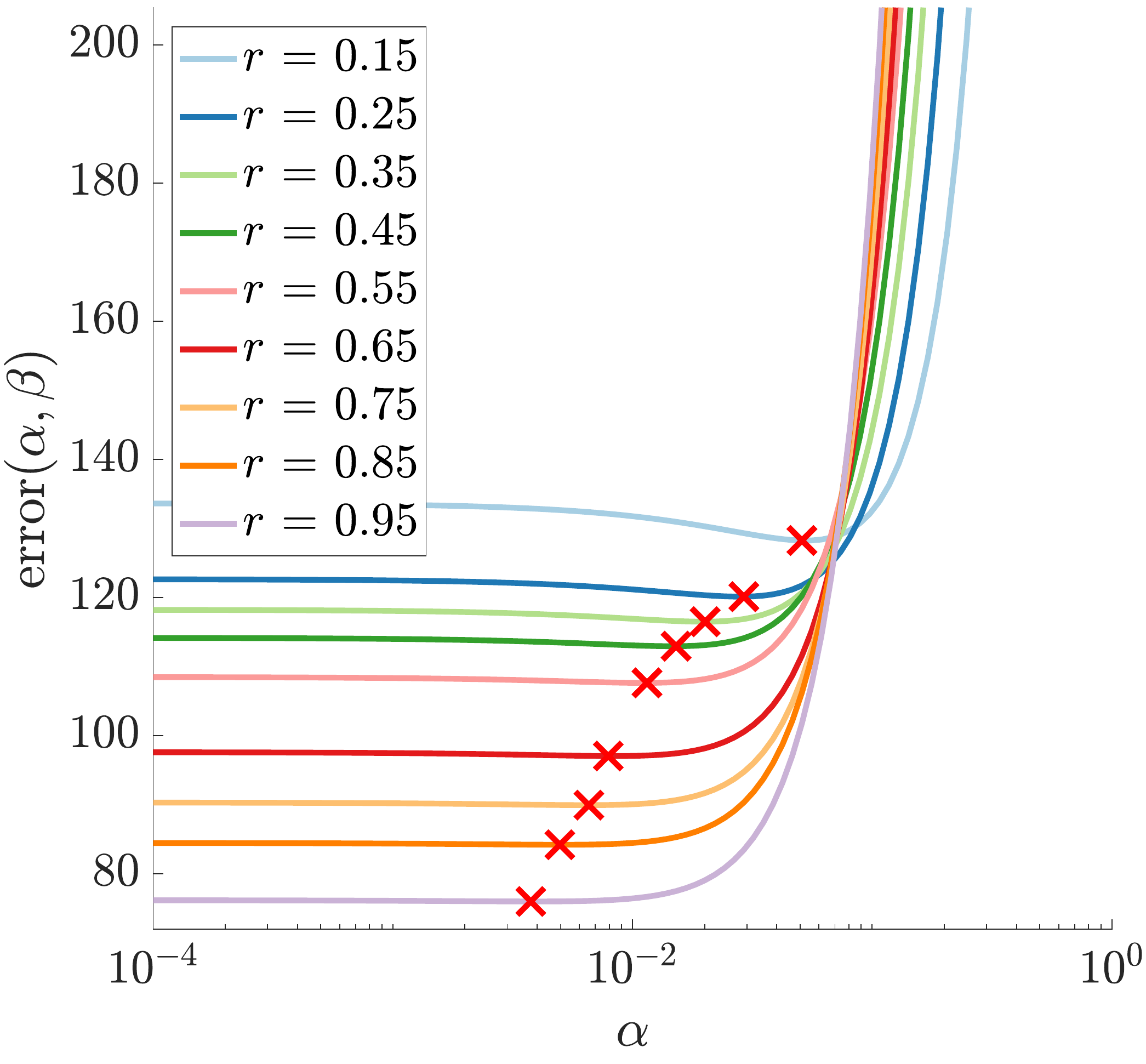}} &
    \subfigure{\includegraphics[ width=0.24\textwidth]{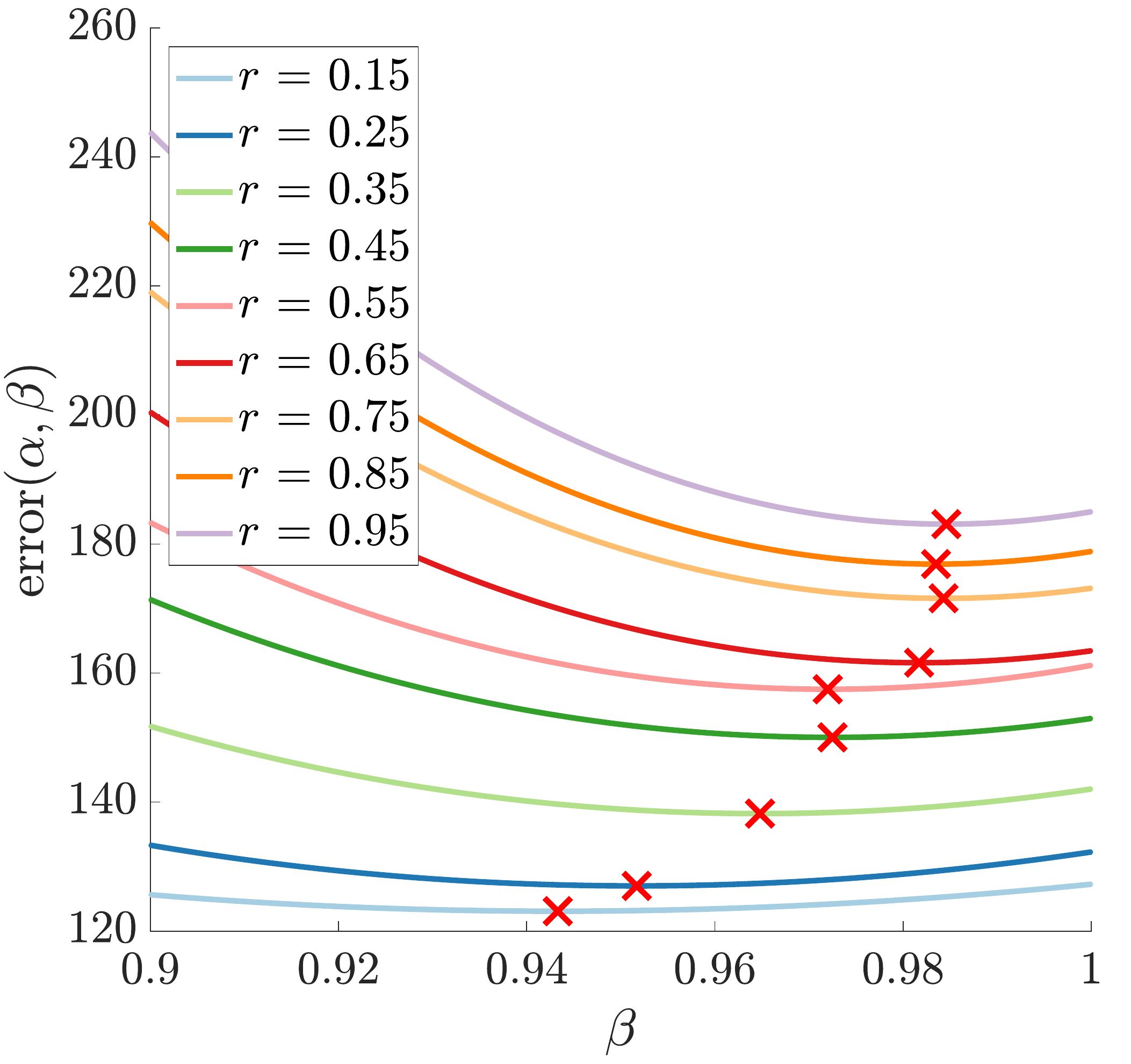}}\\
    (a) & (b) & (c) & (d)
    \end{tabular}
    }
    \vspace{-3mm}
    \caption{\small Experiments on the synthetic data: effect of varying number of trials $n_t$ on the optimum (a) $\alpha$ (fixed $\beta = 1$), and (b) $\beta$ (fixed $\alpha = 0.2$), effect of varying the ratio of compared pairs $r$ on the optimum (c) $\alpha$ (fixed $\beta = 1$), and (d) $\beta$ (fixed $a = 0.25$). The minimum value of each curve is marked with $\bm{\times}$. The error bars are not shown to avoid clutter. Best viewed in color.}
    \label{fig:syn}
    \vspace{-5mm}
    \end{center}
    \end{figure*}
    
The \textit{Rank Centrality} method is based on the Bradley-Terry-Luce (BTL) model for comparative judgment in which a positive weight $\we_i > 0$  is associated with each item $i \in [N]$ such that $\we_i$ reflects the importance (or quality) of the item. Let $\w \in \RR^N_{>0}$ denote the weight vector associated with all the items. For a given pair of distinct items $i, j \in [N]$, the expected probability of preferring $i$ over $j$ is defined as
\begin{equation}
\label{eq:p-gold}
\pe_{ij} \coloneqq \frac{\we_i}{\we_i + \we_j}\, .
\end{equation}
Note that $\we_i > \we_j$ indicates a higher probability of preference of $i$ over $j$. Thus, an ordering based on the actual $\w$ values reflects the optimal expected ranking among the items. Assuming the the pairwise comparison are drawn according to the BTL model, then $\pl_{ij}$ becomes an unbiased estimator of $\pe_{ij}$, i.e. $\pe_{ij} = \EE[\pl_{ij}]$ where the expectation is w.r.t. the set of Bernoulli random variables corresponding to the outcomes of the comparisons.

In order to estimate the weights $\w$ (up to a constant scale), the \textit{Rank Centrality} algorithm utilizes the pairwise comparison information among the pairs of items in $\Pairs$ to construct a Markov chain transition matrix $\P$ where
\[
\Pe_{ij} = 
\begin{cases}
\frac{1}{d_{\max}(i)}\, \pl_{ij} & i \neq j\\
1 - \frac{1}{d_{\max}(i)}\, \sum_{k:\, (i,k) \in \Pairs} \pl_{ik} & i = j
\end{cases}\, ,
\]
in which, $d_{\max}(i)$ denotes the maximum out-degree of node $i$. It has been shown in~\cite{negahban2012} that the stationary distribution of the chain $\P$, denoted by $\s \in \RR^n_{>0}$, approximates the distribution induced by the normalized BTL weights $\w$, that is,
$
\se_i \approx \frac{\we_i}{\sum_j \we_j}\, .
$
Thus, the ordering induced by the stationary distribution $\s$ approximates the ordering induced by the underlying weights $\w$. The results of~\cite{negahban2012} suggest that $\s$ can be used as a proxy for the actual BTL weights $\w$. Thus, we define 
\begin{equation}
\label{eq:p-global}
\pg_{ij} \coloneqq \frac{\se_i}{\se_i + \se_j}\, ,
\end{equation}
Similarly, assuming a BTL model over the items, it can be shown that $\pg_{ij}$ is also an unbiased estimator of $\pe_{ij}$, i.e. $\pe_{ij} = \EE[\pg_{ij}]$, where the expectation is taken over the set of pairwise comparisons $\Pairs$ and the outcomes. As a result, $q^\star_{ij}$ defined in~\eqref{eq:q-star} remains an unbiased estimator of the true expected probabilities $\pe_{ij}$.

\subsection{$\beta$-smoothing}

In many applications such as word embedding, smoothing the estimated probabilities of the items results in an improved performance~\cite{mikolov}. Inspired by these approaches, we replace the global probabilities~\eqref{eq:p-global} with a $\beta$-smoothed version with parameter $\beta \geq 0$ as follows
\begin{equation}
\label{eq:p-beta}
\pg_{ij} \coloneqq \frac{\se^\beta_i}{\se^\beta_i + \se^\beta_j}\, .
\end{equation}
Note that $\beta = 0$ yields $\pg_{ij} = \sfrac{1}{2}$ which corresponds to a uniform distribution (i.e. $\w = \sfrac{1}{N}\,\bm{1}$) over the items. Additionally, $\beta = 1$ corresponds to an identity mapping and values of $\beta > 1$ yield skewed distributions towards popular items.

\section{Experimental Results}
\label{sec:experiments}

In this section we explore the efficacy of the proposed rank-smoothed approach on synthesized as well as real subjective study data.  

\subsection{Synthetic Data}
We investigate the effect of the parameters $(\alpha, \beta)$ on a synthetic dataset in different scenarios. We consider pairwise comparisons on a set of $N = 500$ items for which the BTL weights are drawn from a power-law distribution $P(\we) \propto \we^{\gamma}$ where $\we_{\min} = 0.1$ and $\gamma = 2$. In each experiment, we randomly compare a certain ratio of the total pairs, denoted by $r$, and compare each pair $n_t = n_{ij} + n_{ji}$ times according to the BTL model~\eqref{eq:p-gold}. As the performance measure for the pair of parameters $(\alpha, \beta)$, we report
\[
\error = \sum_{(i,j) \in \Pairs}\, \pe_{ij}\,\log\frac{\pe_{ij}}{q^\star_{ij}} - \pe_{ij} + q^\star_{ij}\, ,
\]
which corresponds to the generalized Kullback–Leibler divergence between the true pairwise probabilities $\pe_{ij}$ in~\eqref{eq:p-gold} and the predicted probabilities $q^\star_{ij}$ given in~\eqref{eq:q-star}. 
We report the average result over $10$ trials. Note that finding the optimal $(\alpha, \beta)$ for a given problem should be considered as a joint optimization. However, we consider a simple case where we fix one parameter and optimize the other. That is, we optimize for $\alpha$ for a fixed value of $\beta$ and vice versa. 

\subsubsection{Effect of number of trials\, $n_t$}

For a fixed ratio of compared pairs $r = 0.15$, we investigate the effect of varying the total number of trials per pair $n_t$ on the parameters $(\alpha, \beta)$. We consider $n_t \in \{3, 5, 10, 20, 50, 100\}$. First, we fix $\beta = 1$ and calculate $\error$ for different values of $\alpha \in [0, 1]$. The result is shown in Fig.~\ref{fig:syn}(a). As can be seen from the figure, the model tends to favor larger values of $\alpha$ (i.e. the empirical probabilities rather than the smoothed ones). This observation is consistent with the fact that the variance of $n_t$ Bernoulli trials goes down as $\sfrac{1}{n_t}$. Thus, the empirical probabilities $\pl_{ij}$ become more accurate estimates of the actual probabilities $\pe_{ij}$.

Next, we fix $\alpha = 0.2$ and vary the value of $\beta$ in the range $[0.5, 1.2]$. The results are shown in Fig.~\ref{fig:syn}(b). Note that the optimal value of $\beta$ approaches $1$ as the number of trials $n_t$ increases. In other words, as the empirical probabilities $\pl_{ij}$ become better estimates of $p_{ij}$, the quality of $\pg_{ij}$ also improves and thus, less smoothing is required.

\subsubsection{Effect of ratio of compared pairs\, $r$}
We also explore the effect of varying the ratio of compared pairs $r$ on the parameters ($\alpha, \beta$). We consider $r \in \{0.15, 0.35, 0.55, 0.75, 0.95\}$. First, we fix $\beta = 1$ and vary the value of $\alpha$ in the range $[0, 1]$. The results are shown in Fig.~\ref{fig:syn}(c). As can be seen from the figure, for larger values of $r$, the stationary distribution $\s$ of the rank-centrality method becomes a better approximate of the actual BTL weights, thus smaller values of $\alpha$ are preferred. 

Next, we fix $\alpha = 0.25$ and vary $\beta$ in the range $[0.9, 1]$. The results are shown in Fig.~\ref{fig:syn}(d). Again, larger values of $r$ results in better estimates of the true probabilities $\pe_{ij}$. Thus, less smoothing is required. Note that for the fix $\alpha = 0.25$, the minimum error value tends to increase with $r$. This is due to the fact that the optimum pair $(\alpha, \beta)$ needs to be jointly optimized for each $r$, rather than fixing $\alpha$ and optimizing~$\beta$.

\begin{figure}[t!]
\vspace{-11 mm}
\begin{center}
\includegraphics*[scale=0.5]{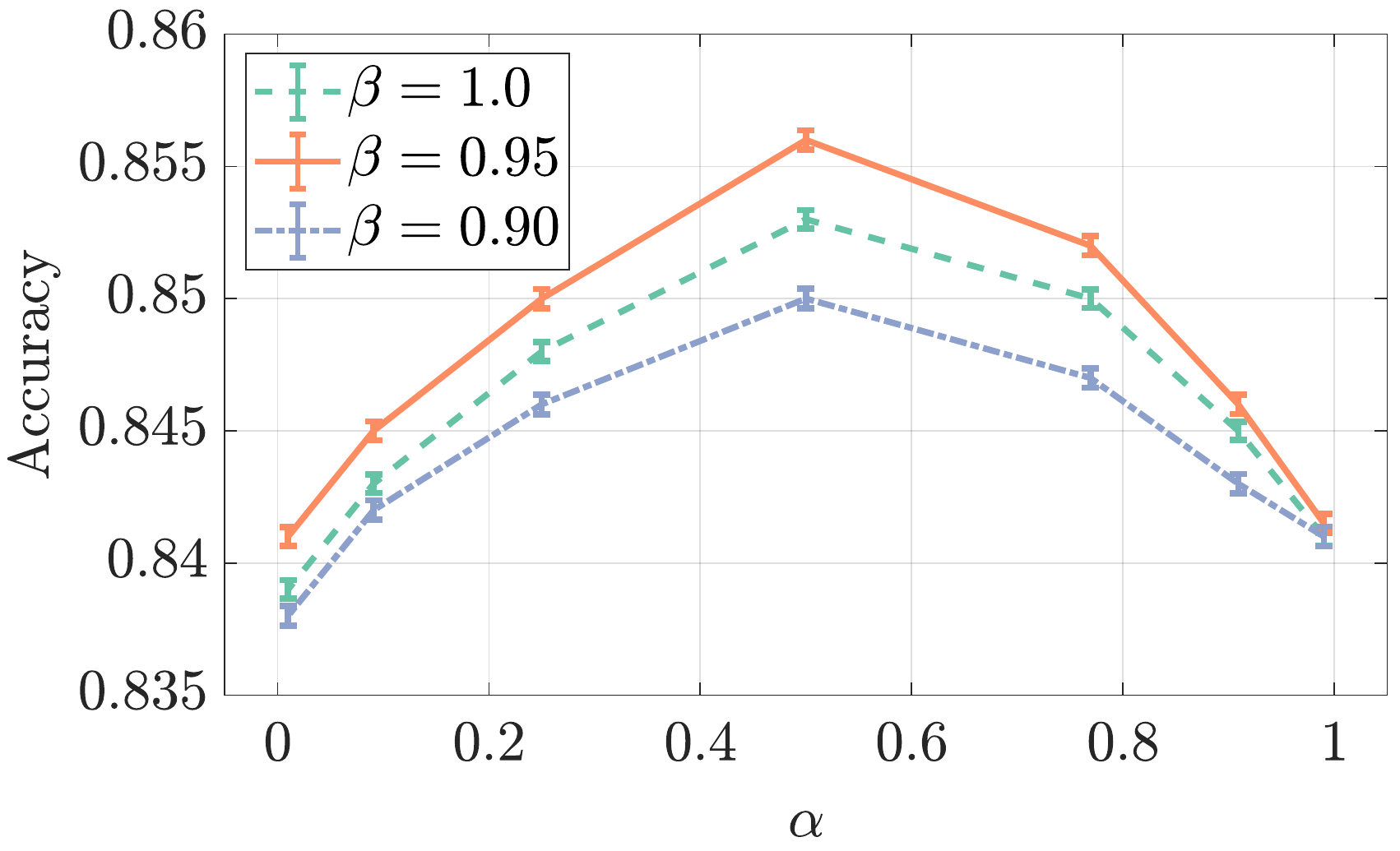}
\end{center}
\vspace{-7 mm}
\caption{\small Accuracy of predicting the majority votes for our subjective study against the parameter $\alpha$ in Eq. \ref{eq:loss-combined}, and various smoothing parameters $\beta$ in Eq. \ref{eq:p-beta}. The accuracy values are averaged for 10 training realizations.} \label{fig:acc}
\vspace{-5 mm}
\end{figure}

\subsection{Subjective Study}

Our dataset consists of a quarter million images donated from Google Photos users. We obtained the necessary permission to use the data in our experimentation, however, we are not allowed to present the image pixels in this paper.

Our subjective study is focused on determining two main qualities of images; sharpness and exposure. In our forced-choice study, we asked raters to ignore image content, and try to compare images based on blurriness and lighting condition. Each image is randomly paired with 24 other images from our dataset. This results in nearly 3 million unique questions. We collected a total of 17 million answers, where each unique question is answered by at least 5 different human raters. Agreement among raters is 51\% for 5 to 0 votes, 26\% for 4 to 1 votes, and 23\% for 3 to 2 votes.


As our CNN model, we use Inception-v2 \cite{szegedy2016rethinking}, and replace its last layer with a spatial pyramid pooling layer \cite{he2015spatial} and a fully connected layer. We initialize the CNN weights from NIMA model~\cite{talebi2018nima}. The weight and bias momentums are set to 0.9, and the learning rate is set to 0.001. Also, after each epoch of training with mini batch of size 128, an exponential decay with decay factor 0.9 is applied to all learning rates. The model is trained for 10 epochs.

We train the CNN by weight sharing with the proposed loss in~\eqref{eq:loss-combined}. Our model is trained on 95\% of the curated dataset, and tested on the remaining pairs. To quantify performance of the model, accuracy of predicting the majority vote for each test pair is measured in Fig.~\ref{fig:acc}. As can be seen, the optimal blending parameter $\alpha$ happens near 0.5. Note that $\alpha=0$ corresponds to relying on the global ranking, and it leads to the lowest accuracy. An interesting observation is that even a small $\alpha$ improves the performance. We also tried various values for the smoothing parameter $\beta$ as in Eq.~\eqref{eq:p-beta}. The best result correspond to $\alpha=0.5$ and $\beta=0.95$.

\begin{figure}[t!]
\vspace{-11 mm}
\begin{center}
\includegraphics*[scale=0.111]{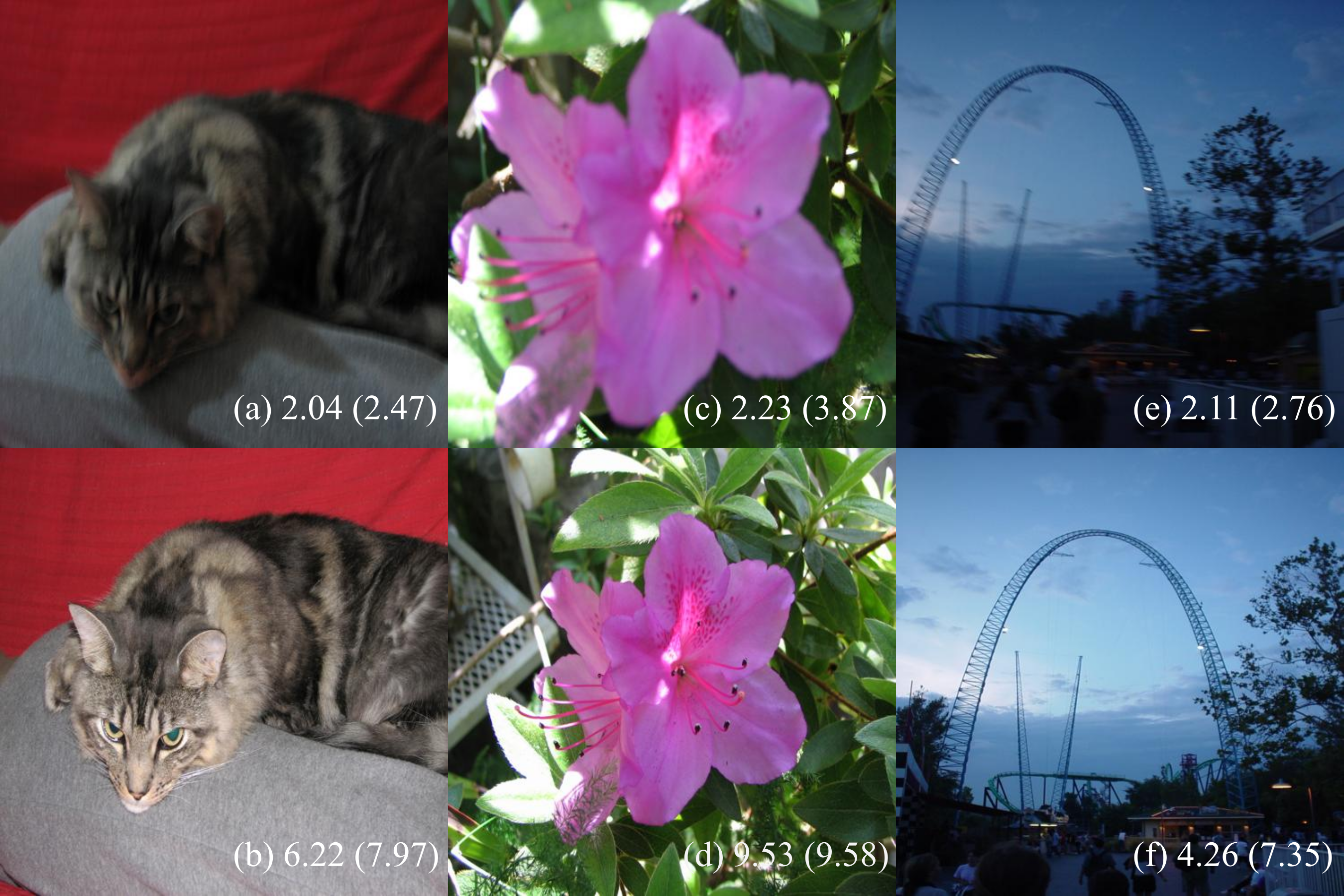}
\end{center}
\vspace{-6 mm}
{\caption{\small Examples from LIVE dataset~\cite{ghadiyaram2015massive}. First score shows our prediction, and score in parenthesis represents the mean raters score from LIVE dataset. Note that higher score means better quality.} \label{fig:live_photos}}
\vspace{-5 mm}
\end{figure}

We present results from our model on the LIVE dataset~\cite{ghadiyaram2015massive} in Fig.~\ref{fig:live_photos}. Although our model is not trained on LIVE dataset, we still obtain a linear correlation of 0.71 with LIVE human ratings. It is worth mentioning that since our data was curated to assess blur and exposure, as it can be seen in Fig.~\ref{fig:live_photos}, our model is appropriately sensitive to blur and exposure changes.

\section{Conclusions}
\label{sec:conclusion}

In this paper we presented a novel approach for learning from pairwise comparisons obtained form subjective studies. Proposed approach does not impose any extra computation at inference, and only requires adjustments in the learning loss. We showed that regularizing the empirical pairwise comparisons with global ranking results in more accurate quality assessment models. Our approach was tested on generic synthesized data, implying that it can be employed beyond the scope of image quality assessment. 

{\small
\bibliographystyle{IEEEbib}
\bibliography{egbib}

}

\end{document}